\documentclass[runningheads]{llncs}
\usepackage{graphicx}
\usepackage{float}
\usepackage{hyperref}

\usepackage[square,sort,comma,numbers]{natbib}
\usepackage{graphicx}
\graphicspath{ {./images/} }
\usepackage{booktabs}
\usepackage{algorithmic}
\usepackage{amsmath}
\DeclareMathOperator*{\argmin}{argmin} 
\usepackage{amssymb}
\usepackage{multirow}
\usepackage[table,xcdraw]{xcolor}
\usepackage[ruled,vlined]{algorithm2e}
\usepackage{multicol,lipsum}
\usepackage{mathtools}
\usepackage{bm}
\usepackage{xcolor}
\setcitestyle{square}
\newcommand\blfootnote[1]{%
  \begingroup
  \renewcommand\thefootnote{}\footnote{#1}%
  \addtocounter{footnote}{-1}%
  \endgroup
}

\begin{document}
\title{Ontology-based $n$-ball Concept Embeddings Informing Few-shot Image Classification}
\titlerunning{Ontology-based Few-shot Image Classification}
%
\author{Mirantha Jayathilaka\orcidID{0000-0002-2462-4833} \and
Tingting Mu\orcidID{0000-0001-6315-3432} \and
Uli Sattler\orcidID{0000-0003-4103-3389}}
\authorrunning{M . Jayathilaka et al.}
%
\institute{Department of Computer Science, The University of Manchester, UK 
\email{\{mirantha.jayathilaka,tingting.mu,uli.sattler\}@manchester.ac.uk}}
\maketitle              
\begin{abstract}
We propose a novel framework named ViOCE that integrates ontology-based background knowledge in the form of $n$-ball concept embeddings into a neural network based vision architecture. The approach consists of two components - converting symbolic knowledge of an ontology into continuous space by learning $n$-ball embeddings that capture properties of subsumption and disjointness, and guiding the training and inference of a vision model using the learnt embeddings. We evaluate ViOCE using the task of few-shot image classification, where it demonstrates superior performance on two standard benchmarks.\blfootnote{Copyright © 2021 for this paper by its authors. Use permitted under Creative Commons License Attribution 4.0 International (CC BY 4.0).} 

\end{abstract}

\section{Introduction}

\begin{figure*}[h!]
\centering
\includegraphics[width=1\textwidth]{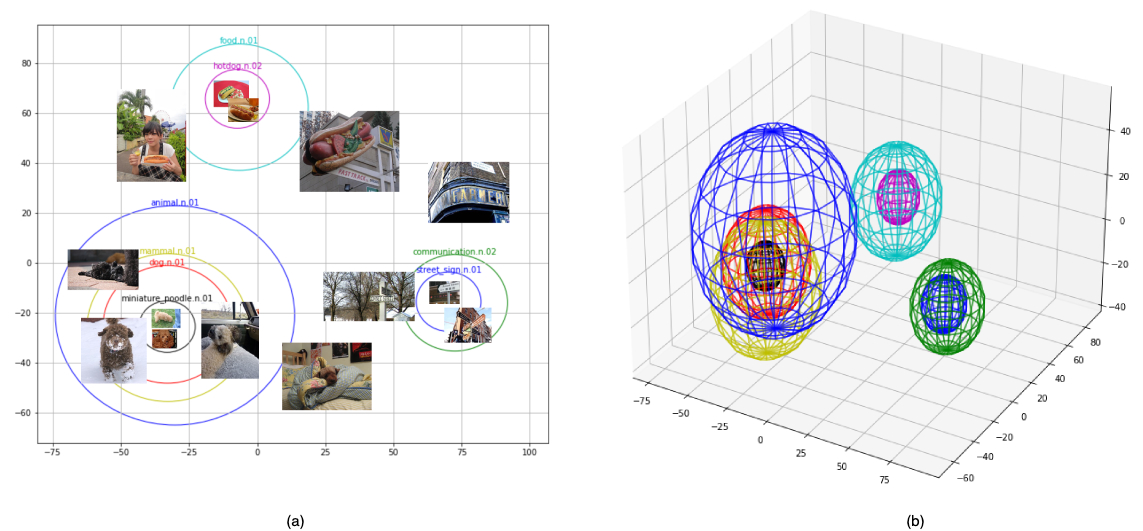}
\caption{The proposed approach classifies images by projecting them towards concept $n$-balls defined in high-dimensional space according to ontology-based background knowledge. (a) shows a snapshot of a few predictions for three miniImageNet\cite{matchingfewshot} classes `miniature poodle', `hotdog' and `street sign' made by a model trained using the ViOCE framework during few-shot image classification. The dimensionality of the $n$-balls is reduced to 2 for visualisation purposes. A correct prediction is an image projected to be inside of the $n$-ball of its ground truth label. Additionally, the surrounding $n$-balls to the ground truth $n$-balls, defined according to the background knowledge, gives us an unique opportunity to measure the `certainty' of the model in classifying each image. For example, not all `miniature poodle' images lie inside the ground truth $n$-ball but an image lying inside `dog' can be identified as semantically meaningful. (b) is a visualisation of the same set of $n$-balls in (a) reduced to a 3-dimensional space in order to provide a clearer idea on the nature of $n$-ball shape and placing.}
\label{fig: prediction glimpse}
\end{figure*}

Ontologies can capture consistent, generalised and structured knowledge that can be used with reasoning tools \cite{hermitreasoner} that ensure knowledge consistency together with the ability to infer new knowledge \cite{alsubait2014measuring}. Sometimes knowledge graphs are also called as ontologies \cite{knowledgegraphvsontologywickramarachchi}, but we identify clear differences. Knowledge graphs tend to be more loosely defined, whereas ontologies have a well-defined semantics that distinguish concepts from the given knowledge specification and other relationships (e.g., hasPart) between concepts bound by logical axioms. Sometimes a knowledge graph can be seen as a specific instantiation of a whole or part of an ontology representing only object-level information \cite{KG3}, whereas ontologies include both concept-level information and objects or terms. 
 
Moreover with powerful reasoning tools, ontologies facilitate the discovery of implicit knowledge from explicitly define knowledge. This study sheds light on the use of ontologies in a machine learning context. We use the Web Ontology Language (OWL) \cite{owlrefernceuli} in constructing our ontologies in this study. 
In order to assess the impact of knowledge integration to a visual recognition task, we chose few-shot image classification \cite{closerlookatfewshot} to be the main task in this study. Few-shot learning in an image classification context focuses on effectively learning the visual features of a class with very few examples. 

The proposed ViOCE framework, we adopt a technique to embed ontology-based knowledge as $n$-balls inspired by the work done by Kulmanov et al. \cite{elembeddings}. 
This embedding can represent specialisations (e.g., Dog SubclassOf Animal) using the property of one $n$-ball enclosing another and partonomies (e.g., Dog hasPart Tail) using translations of $n$-ball positions. In this study, we directly utilise two loss design components of \cite{elembeddings} to capture subsumption and disjointness axioms, while extending their approach with more regularisation components in order to embed large hierarchies in a favourable manner for a downstream vision task. Additionally, we propose the use of the inferred class hierarchy of the input ontology and introduce a technique to evaluate the quality of the learnt embeddings during the embedding learning process. The learnt $n$-ball embeddings can be seen as definitions of space for each concept in consideration that preserves the inferred class hierarchy entailed by the ontology. Next, we introduce a method to use a vision model \cite{frome2013devise, jayathilaka2020visual} to map input images to the space defined by the concept embeddings, informing the vision task with the knowledge captured from the ontology. Figure \ref{fig: prediction glimpse} shows a snapshot of a few predictions for some miniImageNet\cite{matchingfewshot} classes `miniature poodle', `hotdog' and `street sign' made by a model trained using the ViOCE framework. We find that our approach facilitates better transparency on the behaviour of both knowledge embeddings and visual feature learning. 

Overall, we extend \cite{elembeddings} to capture knowledge from an ontology in the form of $n$-ball embeddings and show that they are favourable for the downstream vision task of few-shot image classification. This is also coupled with a technique to measure the quality of the learnt embeddings with respect to the knowledge entailed by the ontology. Next, we propose a technique to utilise the $n$-balls to guide a vision model during its training and inference stages performing few-shot image classification.

\section{Related Work}

An area that inspires the investigation of background knowledge integration in vision is the existing work done in knowledge-based vision systems \cite{knowledgewithvision1, knowledgewithvision2}.  In \cite{knowledgewithvision1}, an interesting categorisation of knowledge that can be used as background knowledge is proposed, namely, permanent theoretical knowledge, circumstantial knowledge, subjective experimental knowledge and data knowledge. Although how these categories are formed is debatable, the importance of looking into different forms of knowledge that can be used as background knowledge is identified. The choice of knowledge form can be very much based on the considered vision application, as pointed out in \cite{knowledgewithvision2}, where the authors curate a number of vision tasks along with the forms of knowledge used to inform the learning process. Out of these, the use of scene graphs, probabilistic ontologies and first-order logic rules grab the attention as promising paths to explore. Investigations into the use of background knowledge in the form of first-Order Logic (FOL) is prominently seen in several studies \cite{hu2016harnessing}. 
 
As shown in \cite{hu2016harnessing}, adaptation of logical knowledge as constraints during the learning process has generated promising results, that reinforces the attempts to use ontologies as background knowledge. The area of neuro-symbolic approaches also provides insights into the use of logical knowledge during the training of artificial neural networks \cite{serafini2016logic}.

In terms of combining other sources of knowledge \cite{mikolov2013efficient} with computer vision, this study is motivated by work such as \cite{frome2013devise, jayathilaka2020visual} and \cite{wang2018zero}, where image features are mapped to a vector space defined by language embeddings. This is identified as informing the image model with more knowledge that do not exist merely in the image features. In the case of \cite{frome2013devise}, the knowledge from an unstructured text corpus is captured in the form of word embeddings to be integrated to the vision architecture. These approaches were mostly evaluated on zero-shot image classification, making use of the distance between points in the vector space defined. These findings motivate the proposed approach in this study, since they allow to extend standard vision models to incorporate language information. In terms of evaluation however, it can be argued that few-shot image classification \cite{qiao2018few} is a better candidate to measure how additional knowledge could help grasp new concepts faster.

In terms of few-shot learning \cite{finn2017model, jayathilaka2019enhancing}, our study is motivated by metric learning methods \cite{vinyals2016matching, qiao2018few} due of their ability to extend standard vision architectures \cite{he2016deep}. These approaches exploits image feature similarities \cite{prototypicalnetsforfewshot} when learning and predicting a vision task.

\section{$n$-Balls and EL Embeddings}
\label{el embeddings preliminary work}

The mathematical concept of ball refers to the volume space bounded by a sphere  and is also called a solid sphere. An $n$-ball usually refers to a ball in an $n$-dimensional Euclidean space. The EL embeddings study \cite{elembeddings} attempts to encode logical axioms by positioning $n$-balls. We explain  how it works for encoding subsumption   and disjointness  as they are the most relevant  to our work.  Each concept $P$ is embedded as an $n$-ball with its centre denoted by $\bm c_{P} \in \mathbb{R}^{n}$ and the radius  by $r_{P} \in \mathbb{R}$. The basic idea is to  move one ball inside the other for subsumption and to push two balls to stay away for disjointness. 
The following loss is minimized to encode $\mathcal{O} \models P \sqsubseteq Q$: 
\begin{equation}
\label{eq: loss1}
\begin{split}
l_{P \sqsubseteq Q} & (\bm c_{P}, \bm c_{Q}, r_{P}, r_{Q}) \\
= &  \max(0, \|\bm c_{P} - \bm c_{Q}\|_2 + r_{P} - r_{Q} - \gamma) \\
    & + \big| \|\bm c_{P}\|_2 - 1 \big| + \big| \|\bm c_{Q}\|_2 - 1\big|,
\end{split}
\end{equation}
where $\|\cdot\|_2$ denotes the $l_2$  norm and $\gamma \in \mathbb{R}$ is a user-set hyperparameter.  It enforces the inequality   $\|\bm c_{P} - \bm c_{Q}\|_2 \leq r_{Q} - r_{P} +\gamma$, meanwhile regulates the  ball centers to be close to a unit sphere. Through controlling the sign  of $\gamma$, the user can adjust whether  to push the $P$ ball completely inside the $Q$ ball. In a similar fashion, the loss for encoding $\mathcal{O} \models P \sqcap Q \sqsubseteq \bot$ is given as
\begin{equation}
\label{eq: loss2}
\begin{split}
   l_{P \sqcap Q \sqsubseteq \perp} & (\bm c_{P}, \bm c_{Q}, r_{P}, r_{Q})  \\
 =&  \max (0, - \|\bm c_{P} - \bm c_{Q}\|_2+ r_{P} + r_{Q} + \gamma) \\ 
    & + \big| \|\bm c_{P}\|_2 - 1 \big| + \big| \|\bm c_{Q}\|_2 - 1 \big|.
\end{split}
\end{equation}
It enforces the inequality   $\|\bm c_{P} - \bm c_{Q}\|_2 \geq r_{Q} + r_{P} +\gamma$. According to the setting of $\gamma$, the user can decide how far the two balls are pushed away.

\section{Proposed Method: ViOCE}
\label{vioce}

We study how to effectively integrate ontology-based background knowledge to improve few-shot image classification. More specifically, this paper is focused on using additional hierarchical knowledge about the different classes to help image classification,  achieving reduced   data dependency of vision model architectures that are based on deep neural networks.

Adopting few-shot image classification as our benchmark \cite{leveragingfewshot}, we train a neural vision model using a set of  background images $BI = \{(\bm I_i, y_i)\}_{i=1}^m$ (base set) from $\mathcal{K}$ classes with $y_i \in C_B=\{c_1,  c_2, \ldots c_\mathcal{K}\}$ and a set of few-shot images $FI = \{(\bm I_i, y_i)\}_{i=1}^s$ (novel set) from $w$ classes with $y_i \in C_F=\{\tilde{c}_1,  \tilde{c}_2,  \ldots, \tilde{c}_w\}$, where $C_B \cap C_F = \emptyset$, and $\bm I_i$ denotes the raw image vectors containing pixel values.   The few-shot success  is usually assessed  by how accurate a model can select a correct class from  the candidate class set $C_F $ for a new image from the few-shot classes. This is often referred to as the $w$-way $s$-shot few-shot image classification.   We construct an ontology $\mathcal{O}$ by using the class label information $C_B $ and $C_F$,  and also WordNet.  It provides information on relationships that can exist among the class labels, containing knowledge regarding to ``SubClassOf" and  ``DisjointClasses".  These  define the subsumption and disjointness  axioms in the ontology.

\begin{figure*}[!h]
\centering

\includegraphics[width=1\textwidth]{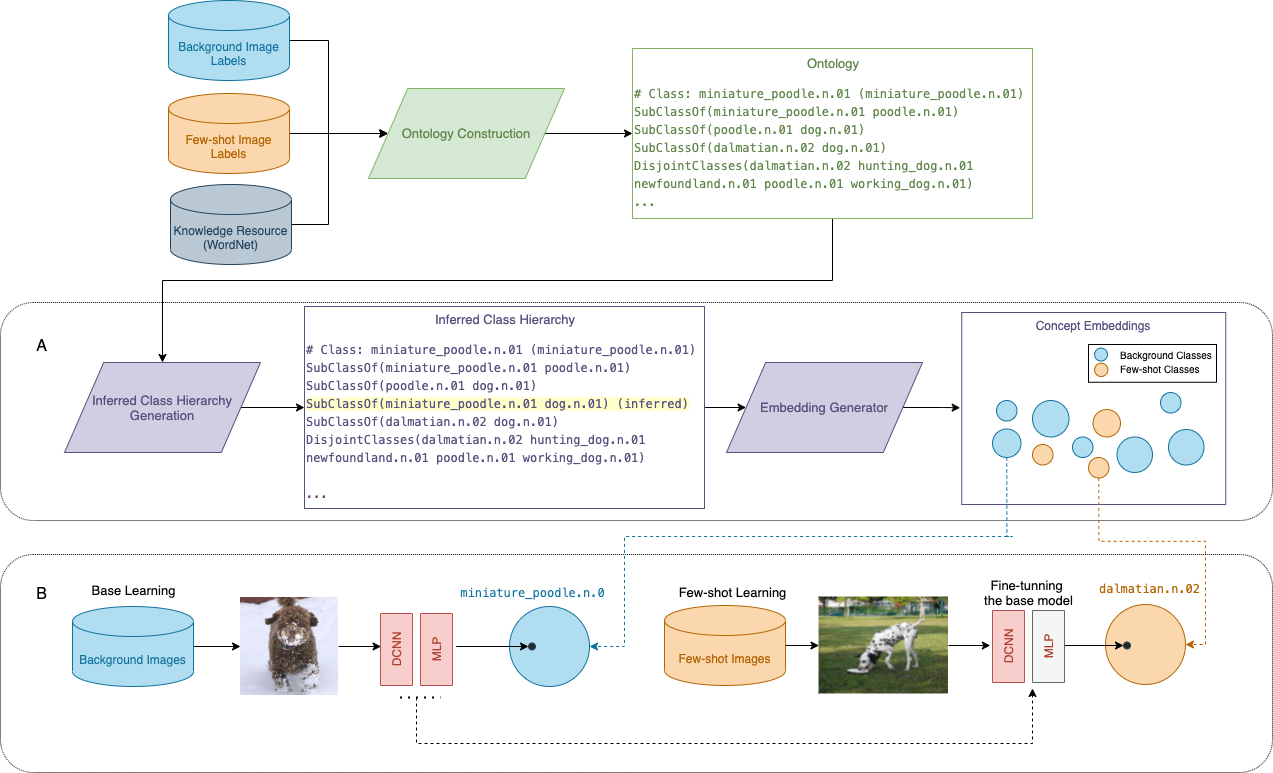}
\caption{The overview of the proposed ViOCE framework. If a suitable ontology for the task does not exist, the approach starts from constructing an ontology for the image labels capturing relationships between them based on an external knowledge resource (in our case WordNet). Subsequently the approach follows the two main components of the framework - A) Concept embedding learning process that starts with computing the inferred class hierarchy ($ICH$) of the input ontology and then generates $n$-ball embeddings for all the concepts found in the ontology. B) Visual model (DCNN+MLP) training where, first the background images are used to train a base model which gets fine-tuned (only MLP) using the few-shot images to produce the final model. During both base learning and few-shot learning processes, the concept embeddings guide the learning process by setting the objective of the model to project the image feature points inside the correct $n$-ball representing the ground truth label of an input image. }
\label{fig: approach design}
\end{figure*}

We propose ViOCE as a framework to improve few-shot image classification by integrating information provided by $\mathcal{O}$, $BI$ and $FI$. It is composed of two main components:   (1) to embed classes in $C_B$ and $C_F$ as $n$-balls  based on the constructed $\mathcal{O}$,    (2) to embed images in the same Euclidean space as the $n$-balls with a suitable arrangement, and to infer the class for a query image based on its image embedding and the $n$-ball embeddings of the candidate classes.  Figure \ref{fig: approach design} shows the general framework flow with an overview of all processes and data inputs. 

\vspace{-1cm}

\subsection{Concept $n$-Ball Embeddings }
\label{subsec: concept embedding learning}
\vspace{-1cm}
 We build  upon  the EL embedding technique \cite{elembeddings}  to learn a set of $n$-balls for all concepts $\widetilde{\mathcal{O}}$ in the ontology $\mathcal{O}$, which is referred to  as a \textit{concept embedding}.  We extract subsumption  and disjointness axioms to define the class hierarchy of  the ontology $\mathcal{O}$.   It has been noticed that the entailed transitive relations such as if \textit{Poodle SubclassOf Dog} and \textit{Dog SubclassOf Animal}, then \textit{Poodle SubclassOf Animal} are usually not well reflected by the learned $n$-balls. To overcome this, we use the inferred class hierarchy (ICH). Assuming all the concepts are satisfiable with $\mathcal{O}$, the ICH is computed according to Equation \ref{eq: inferred class hierarchy equation}. ICH contains all possible subsumption relations according to the definition of $\mathcal{O}$.  
\begin{equation}
\label{eq: inferred class hierarchy equation}
  \textmd{ICH}(\mathcal{O}) = \{ P \sqsubseteq Q | P \ne Q, P, Q \in \widetilde{\mathcal{O}}, \mathcal{O} \models P \sqsubseteq Q \}.   
\end{equation}

\vspace{-0.5cm}

If simply to follow Eqs. (\ref{eq: loss1}) and (\ref{eq: loss2}), the radius of the learned $n$-ball for a leaf concept, which corresponds to an image class in $C_B$ or $C_F$, can end up being very small,  in order to fit into the balls of its ancestor concepts. Since in the image embedding learning, we  will  map  each  image as a data point inside the $n$-ball corresponding to its ground truth class,  an overly small radius can affect the learning accuracy. To tackle this, we introduce a  regularisation term in Eq. (\ref{eq: loss3}) to prevent radius shrinkage.
Also,  the embedding quality can deteriorate as the class hierarchy of the input ontology becomes larger. To improve the  embedding quality, we introduce an extra hyperparameter in Eq. (\ref{eq: lossF}) to explore potentially more expressive design spaces,  which is  supported by an additional parameter tuning process. Finally, we  minimise the following loss function:
\begin{align}
\label{eq: lossF}
l_c & \left( \{\bm c_P\}_{P\in \widetilde{\mathcal{O}}} , \{r_P\}_{P\in \widetilde{\mathcal{O}}}\right) \\
\nonumber
=& \sum_{\textmd{ICH}(\mathcal{O}) \models P \sqsubseteq Q  }  \max(0, \|\bm c_{P} - \bm c_{Q}\|_2 + r_{P} - r_{Q} - \gamma) \\ 
\nonumber
   & + \sum_{\mathcal{O} \models C \sqcap D \sqsubseteq \bot} \max (0, - \|\bm c_{P} - \bm c_{Q}\|_2+ r_{P} + r_{Q} + \gamma)  \\ 
  \label{eq: loss3}
   &+   \sum_{P\in \widetilde{\mathcal{O}}} \max (0, \psi\sqrt{N_h - L(P)} - r_P)\\
     \label{eq: loss4}
      & + \sum_{P\in \widetilde{\mathcal{O}}} N(P)\big| \|\bm c_{P}\|_2 - \phi \big|
\end{align}
Here, $N_h$ denotes the total level number contained by the class hierarchy, and $L(P)$ denotes the level of the concept $P$ in the hierarchy, e.g.,, the top-most concept has level 1. $N(P)$ denotes the number of times the concept $P$ appears in the extracted  axioms.  Both $\psi, \phi  >0$   are hyperparameters. Eq. (\ref{eq: loss3}) restrict the radius of the concept $P$ 's $n$-ball to be no less than  $\psi\sqrt{N_h - L(P)}$. The top-level concepts are allowed to have larger $n$-balls than the bottom ones.

\vspace{-0.3cm}

\subsection{Hyperparameter Tuning of $n$-ball Embeddings}
\label{subsubsec: evaluation of CE}

Three parameter tuning scores are proposed by examining whether  $\|\bm c_P - \bm c_Q\| \leq  r_Q - r_P $ holds for a ground truth subsumption $ICH(\mathcal{O}) \models P \sqsubseteq Q$. All the ground truth subsumptions are considered as positive instances.  If the inequality holds, it is considered as a positive prediction.   The classical $F_1$ score, which is the harmonic mean of the precision and recall, is used to assess the prediction accuracy  of these subsumptions.   We calculate two versions of $F_1$ score, one is referred to as $F_1^{(\textmd{all})}$ based on all the subsumptions extracted from $ICH(\mathcal{O})$. The other only considers the subsumptions involving the leaf concepts, which correspond  to all the classes in $C_B$ and $C_F$,  as well as their direct parent classes. This score is referred to as $F_1^{(\textmd{leaf})}$.
The third parameter tuning score $S_D$ examines the disjointness between the leaf concepts. Enumerating all the pairs of leaf concepts, $S_D$ is equal to the number of pairs for which the condition $\|\bm c_P - \bm c_Q\| \geq  r_P + r_Q $ holds. A higher $S_D$ indicates less overlapping between the $n$-balls of the leaf concepts. We need $S_D$ to be greater that a threshold value of $\mathcal{T}$. 

A good concept embedding result should have high $F_1^{(\textmd{all})}$, $F_1^{(\textmd{leaf})}$ and $S_D$ scores.
We compute these scores as a mandatory step at the end of each embedding learning process. The hyperparameters governing the scores are $\gamma$, $\phi$ and $\psi$. We use grid search to find the best combination of these parameters that would result in the best $F_1^{(\textmd{all})}$, $F_1^{(\textmd{leaf})}$ and $S_D$ scores.

\vspace{-0.2cm}
\subsection{Image Embedding Learning}
\label{sec: ViOCE training and inference}
Our vision model is composed of a base DCNN architecture coupled with a multi-layer perceptron (MLP).  The DCNN computes the visual features for an image by taking its raw pixel representation vector as the input: $\bm  f_i = \phi_{\textmd{D}}(\bm I_i, \bm\theta_{\textmd{D}})$ where $\bm  f_i \in \mathbb{R}^d$. The MLP is responsible for mapping the visual features $\bm  f_i$  to the $n$-dimensional Euclidean space where the $n$-ball concept embeddings sit: $\bm  h_i = \phi_{\textmd{M}}(\bm f_i,\bm\theta_{\textmd{M}})$ where $\bm  h_i \in \mathbb{R}^n$. We use $\bm\theta_{\textmd{D}}$ and $\bm\theta_{\textmd{M}}$ to denote the neural network parameters to be trained for the DCNN and MLP, respectively.
The idea is to identify visual features of an image (using a DCNN) so that they can be mapped (by an MLP) as a data point inside the $n$-ball of its ground truth class. For example, an image containing the visual features of a ``poodle" should be mapped inside the $n$-ball of the ``poodle"  concept learnt from the ontology. 

To achieve  this, the following a pairwise ranking loss is used to optimise the network parameters:
\begin{equation}
\label{eq: vision model loss}
\resizebox{.9 \textwidth}{!}{%
 $ l_I (\bm{\theta_{D}}, \bm{\theta_{M}}) = \sum_{i=1}^{m} \bigg[\max (0, \|\bm{c}_{P} - \bm{h}_i \|_2 - \mu r_{P} )+\sum_{Q\in C_i^{(-)}} \max(0, \nu r_{Q} -\| \bm{c}_Q - \bm{h}_i \|_{2}) \bigg], $
}
\end{equation}
where $\mu, \nu >0$ are hyperparameters. The set $C_i^{(-)}$ contains the negative classes defined for each image $\bm I_i$ of the positive class with its embedding computed by $\bm h_i =\phi_{\textmd{M}}(\phi_{\textmd{D}}(\bm I_i, \bm\theta_{\textmd{D}}),\bm\theta_{\textmd{M}})$. When setting $\mu=\nu=1$, the loss enforces  $\left\|\bm c_{P} -  \bm h_i  \right \|_2 \leq r_{P}$, pushing the embedded image point to stay inside the $n$-ball of the correct concept class $P$,  while $\left\|\bm c_{Q} -  \bm h_i  \right \|_2 \geq r_{Q}$, to stay outside the $n$-ball  of the incorrect concept class $Q$. The hyperparameters $\mu$ and $\nu$ are placed to  control the intensity of this effect, e.g.,, $\mu < 1$ requiring  to lie closer to the center which makes the task  harder.

A  specification  crucial to learning performance is the selection of negatives concepts in $C_i^{(-)}$. Following the notion of  ``hard negatives" in \cite{muhardnegatives}, we  select ``hard negatives" for each positive concepts based on similarity.  For example, the ``poodle" concept is more similar to ``golden retriever" in contrast to the ``street sign", therefore it is more challenging to distinguish between  ``poodle"  and ``golden retriever".  So we choose as the hard negatives the more similar concepts to a positive concept. Specifically, we evaluate similarities between concepts by Euclidean distances between the centre vectors of their corresponding $n$-balls, and perform  k-means clustering based on these.

After clustering the centre vectors of the leaf concepts (image classes),  for each image class, all the other image classes from the same cluster as it are treated as the ``hard negatives" and are included to $C_i^{(-)}$.

In practice, we first train the DCNN and MLP from scratch by minimising Eq. (\ref{eq: vision model loss}) using the background images $BI$. This is called base learning (BL).
Then, we fine tune the MLP by using the few-shot images $FI$ by minimising the same loss, but keep the weights of  DCNN fixed. This called the few-shot learning (FSL).

We test the vision model using the testing images of $FI$ ($FI_{te}$) after the fine-tuning of MLP in the $FSL$ stage. During inference, a prediction is made by finding the $n$-ball which an image feature projection lies in. Let $\mathcal{U} = \| \bm c_P - \bm h \| - r_P$, where $\bm h$ is an output feature for a query image from the vision model and $\bm c_P$ and $r_P$ are the centre and radius of a selected $n$-ball of $P$ respectively. If $\mathcal{U} \leq 0$, we find that $\bm h$ lies inside the $n$-ball of $P$. Hence the classification of $\bm h$ will be class $P$. In case some $\bm h$ does not lie inside any of the $n$-balls of the $w$ classes in the few-shot task, we choose the closest lying $n$-ball centre $\bm c_i$ out of the classes to $\bm h$, where $\argmin_{\bm c_i (i = 1, 2,..,w)} \big( \| \bm c_i - \bm h \| \big)$, as the prediction. The proportion of the correct predictions out of all images in $FI_{te}$ is recorded as the accuracy of the vision model in this study.

\section{Experiment Setting}

MiniImageNet dataset consists of 60,000 images of 100 classes from ImageNet where each class carries 600 example images \cite{matchingfewshot}. Following the same splitting as in \cite{spatialfewshot}, 80 and 20 classes were allocated for training and testing respectively. 

TieredImageNet dataset is larger in size than miniImageNet, containing 608 classes from ImageNet \cite{tieredimagenet}. Its classes are acquired based on 34 higher-level categories. We use a training set consisting of 26 higher-level categories with 448 classes, and testing set of 8 higher-level categories with 160 classes.

We construct two new ontologies based on the image labels of the datasets for each few-shot image classification benchmark. All selected datasets are subsets of ImageNet \cite{imagenet_cvpr09}, where WordNet \cite{miller1995wordnet} synsets are used to annotate all images. This offered the opportunity to use the information from WordNet to formulate more knowledge about the image labels. 
 
We chose the hypernym tree of WordNet to be the source of the class hierarchy in this study, where given a label, the corresponding synset name together with all other synsets above it until the root (\textit{entity.n.01}) was extracted. All these concepts were included in the ontology\footnote{The used ontologies can be accessed via \url{https://github.com/miranthajayatilake/ViOCE-Ontologies}}. The dimensionality of the concept embeddings was chosen to be 300. During all experiments, ResNet50 \cite{ResNet50} architecture was chosen to be the base network and the MLP was composed of 5 layers with sizes of 2048, 1024, 512, 512 and 300.

\section{Results}

\subsection{Few-shot image classification results}
\label{subsec: few-shot results}

ViOCE is evaluated by comparing with the performance of several existing approaches according to \cite{tian2020rethinkingfewshot} under the same configuration. We conduct experiments for $w=\{5,20\}$ and $s=\{1,5\}$.
 
Table \ref{tab:few-shot results} reports the 5-way 1-shot and 5-shot performance comparisons. It can be seen that ViOCE surpasses the the performance of all other approaches in every 5-way tasks with both datasets, while achieving $>$90\% accuracy in miniImageNet 5-shot task.

\begin{table*}[!h]
\centering
\caption{5-way 1-shot and 5-shot accuracy comparison with existing approaches using miniImageNet and tieredImageNet benchmarks. All accuracies are reported with 95\% confidence intervals. }
\label{tab:few-shot results}
\resizebox{\textwidth}{!}{%
\begin{tabular}{@{}lcccc@{}}
\toprule
\multicolumn{1}{c}{}                                 & \multicolumn{2}{c}{\textbf{miniImageNet 5-way}}                                                                                                       & \multicolumn{2}{c}{\textbf{tieredImageNet 5-way}}                                                                                                     \\ \cmidrule(l){2-5} 
\multicolumn{1}{c}{\multirow{-2}{*}{\textbf{Model}}} & \textbf{1-shot (\%)}                                                      & \textbf{5-shot (\%)}                                                      & \textbf{1-shot (\%)}                                                      & \textbf{5-shot (\%)}                                                      \\ \cmidrule(r){1-1}
MAML (Finn et al.)                                                 & 48.70 $\pm$ 1.84                                             & 63.11 $\pm$ 0.92                                             & 51.67 $\pm$ 1.81                                             & 70.30 $\pm$ 1.75                                             \\
Matching Networks (Vinyals et al.)                                    & 43.56 $\pm$ 0.84                                             & 55.31 $\pm$ 0.73                                             & -                                                                         & -                                                                         \\
IMP (Allen et al.)                                                  & 49.20 $\pm$ 0.70                                             & 64.7 $\pm$ 0.70                                              & -                                                                         & -                                                                         \\
Prototypical Networks (Snell et al.)                                & 49.42 $\pm$ 0.78                                             & 68.20 $\pm$ 0.66                                             & 53.31 $\pm$ 0.89                                             & 72.69 $\pm$ 0.74                                             \\
Relational Networks (Sung et al.)                                 & 50.44 $\pm$ 0.82                                             & 65.32 $\pm$ 0.70                                             & 54.48 $\pm$ 0.93                                             & 71.32 $\pm$ 0.78                                             \\
AdaResNet (Munkhdalai et al.)                                            & 56.88 $\pm$ 0.62                                             & 71.94 $\pm$ 0.57                                             & -                                                                         & -                                                                         \\
TADAM (Oreshkin et al.)                                               & 58.50 $\pm$ 0.30                                             & 76.70 $\pm$ 0.30                                             & -                                                                         & -                                                                         \\
Shot-Free (Ravichandran et al.)                                           & 59.04 $\pm$ n/a                                              & 77.64 $\pm$ n/a                                              & 63.52 $\pm$ n/a                                              & 82.59 $\pm$ n/a                                              \\
MetaOptNet (Lee et al.)                                          & 62.64 $\pm$ 0.61                                             & 78.63 $\pm$ 0.46                                             & 65.99 $\pm$ 0.72                                             & 81.56 $\pm$ 0.53                                             \\
Fine-tuning (Dhillon et al.)                                         & 57.73 $\pm$ 0.62                                             & 78.17 $\pm$ 0.49                                             & 66.58 $\pm$ 0.70                                             & 85.55 $\pm$ 0.48                                             \\
LEO-trainval (Rusu et al.)                                        & 61.76 $\pm$ 0.08                                             & 77.59 $\pm$ 0.12                                             & 66.33 $\pm$ 0.05                                             & 81.44 $\pm$ 0.09                                             \\
Embedding-distill (Tian et al.)                                   & 64.82 $\pm$ 0.60                                             & 82.14 $\pm$ 0.43                                             & 71.52 $\pm$ 0.69                                             & 86.03 $\pm$ 0.49                                             \\
ViOCE                                        & \multicolumn{1}{r}{\textbf{65.71 $\pm$ 0.13}}                & \multicolumn{1}{r}{\textbf{93.65 $\pm$ 0.07}}                & \multicolumn{1}{r}{\textbf{73.4 $\pm$ 0.13}}                 & \multicolumn{1}{r}{\textbf{88.95 $\pm$ 0.09}}                \\ \bottomrule
\end{tabular}%
}
\end{table*}

The study further extends the evaluation with the miniImageNet dataset to the task of 20-way 1-shot and 5-shot classification. In this case, considering all the 20 few-shot classes offers a bigger challenge to the model, having to distinguish between more classes with a few examples. Table \ref{tab:few-shot resluts 20 way} presents the result comparison on this task. ViOCE surpasses the performance of existing approaches in both 1-shot and 5-shot tasks with comfortable margins.

\begin{table}[!h]
\centering
\caption{20-way 1-shot and 5-shot accuracy comparison with existing approaches using miniImageNet dataset. }
\label{tab:few-shot resluts 20 way}
\begin{tabular}{@{}lcc@{}}
\toprule
\multicolumn{1}{c}{}                                 & \multicolumn{2}{c}{\textbf{miniImageNet 20-way}}     \\ \cmidrule(l){2-3} 
\multicolumn{1}{c}{\multirow{-2}{*}{\textbf{Model}}} & \textbf{1-shot (\%)}          & \textbf{5-shot (\%)} \\ \cmidrule(r){1-1}
MAML (Finn et al.)                                                & 16.49                         & 19.29                \\
Meta LSTM (Ravi et al.)                                            & 16.70                         & 22.69                \\
Matching Networks (Vinyals et al.)                                   & 17.31                         & 26.06                \\
Meta SGD (Li et al.)                                            & 17.56                         & 28.92                \\
Deep Comparison Network (Zhang et al.)                             & 32.07                         & 47.31                \\
TIM-GD (Boudiaf et al.)                                              & 39.30                         & 59.50                \\
ViOCE                                        & \textbf{48.02}                & \textbf{84.13}       \\ \bottomrule
\end{tabular}%
\end{table}

Another interesting observation during the $BL$ stage of ViOCE was the behaviour of the training and testing accuracies of the vision model. With miniImageNet for example, the model was trained with 500 images per class across 80 classes, which is comparable to a standard image classification task. The training and testing accuracies were 85.32\% and 95.36\% respectively. The higher testing accuracy demonstrates the better generalisation ability of the learnt model. We argue that this effect is due to not forcing the image features to a fixed point as done in a standard training setting. The $n$-ball embeddings define a volume of space for each class providing more flexibility for the arrangement of image feature points. 

\section{Conclusion}
\vspace{-0.2cm}
We show that the introduction of ontology-based background knowledge to a visual model can improve its performance in the task of few-shot image classification. 

The proposed ViOCE framework is capable of utilising the $n$-ball concept embeddings in an effective way to inform the training and inference procedures of a vision model, and producing superior performance on two benchmarks. 
 
In future, we plan to extend this study to evaluate the semantically meaningful errors in classification and utilise multi-relational knowledge when learning concept embeddings.

\bibliographystyle{splncs04}
\bibliography{references}

\end{document}